# Recognition of Indian Sign Language in Live Video

Joyeeta Singha
Department of Electronics and Communication
DBCET, Assam Don Bosco University
Guwahati, Assam

Karen Das
Department of Electronics and Communication
DBCET, Assam Don Bosco University
Guwahati, Assam

## ABSTRACT
Sign Language Recognition has emerged as one of the important area of research in Computer Vision. The difficulty faced by the researchers is that the instances of signs vary with both motion and appearance. Thus, in this paper a novel approach for recognizing various alphabets of Indian Sign Language is proposed where continuous video sequences of the signs have been considered. The proposed system comprises of three stages: Preprocessing stage, Feature Extraction and Classification. Preprocessing stage includes skin filtering, histogram matching. Eigen values and Eigen Vectors were considered for feature extraction stage and finally Eigen value weighted Euclidean distance is used to recognize the sign. It deals with bare hands, thus allowing the user to interact with the system in natural way. We have considered 24 different alphabets in the video sequences and attained a success rate of 96.25%.

## Keywords
Indian Sign Language (ISL), Skin Filtering, Eigen value, Eigen vector, Euclidean Distance (ED), Computer Vision.

## 1. INTRODUCTION
A Sign Language is a language in which communication between people are made by visually transmitting the sign patterns to express the meaning. It is a replacement of speech for hearing and speech impaired people. Thus, because of which it has attracted many researchers in this field from long. Many researchers have been working in different sign languages like American Sign Language, British Sign Language, Taiwanese Sign Language, etc. but few works has been made progress on Indian Sign Language.

The hearing impaired people becomes neglected from the society because the normal people never try to learn ISL nor try to interact with the hearing impaired people. This becomes a curse for them and so they mostly remain uneducated and isolated. Thus recognition of sign language was introduced which has not only been important from engineering point of view but also for the impact on society.

Our paper aims to bridge the gap between us and the hearing impaired people by introducing an inexpensive Sign Language Recognition technique which will allow the user to understand the meaning of the sign without the help of any expert translator. Computers are used in communication path which helps in capturing of the signs, processing it and finally recognizing the sign.

Several techniques have been used by different researchers for recognizing sign languages or different hand gestures. Some researchers worked with static hand gestures, while some worked with video and real time. Researchers in [1-4][11][12] worked with static images. In our previous paper [1], Karhunen-Loeve Transform was used for recognition of different signs but was limited to gestures of only single hand. Accuracy rate obtained was 96%. Bhuyan [2] achieved a success rate of 93% in his paper where he used Homogenous Texture Descriptors to calculate the inflexive positions of fingers and abduction angle variations were also considered. Features in [3] were extracted using Gabor filter and PCA and ANN used for recognition of the Ethiopian Sign Language with a high success rate of 98.5%. Ghotkar [11] in his paper used Camshift algorithm for tracking of hand and for features, Haudsoff Distance and Fourier Descriptor were considered. Recognition was achieved using Genetic Algorithm. In paper [4], Indian sign language was recognized using Eigen value weighted Euclidean distance based classifier with an accuracy rate of 97%. It removed the difficulty faced by [1-3][11][12] for gestures using both hands.

Many research works [5-10] has been done with the video and real time. Chou [5] used HMM for recognition of hand gestures consisting of both hands with an accuracy rate of 94%. Neural Network based features and Hidden Markov Model was used in [6] for recognizing various hand gestures in video. Starner in [7] used Hidden Markov Model for recognition of American Sign Language and achieved a success rate of 99% his work was limited to colored gloves. In [8] skin filtering, moment invariants based features along withANN was used for recognition of different gestures with a success rate of 92.85%. [9][10] were works done to recognize Taiwanese Sign. [9] usedHidden Markov Model in real time. But their work was limited to use of data gloves and recognition of single hand gestures with a low accuracy rate of 84%.In [10] the same difficulty of using colored gloves was present but both static and dynamic hand gestures could be recognized using Support vector machines and Hidden Markov Model.

Thus we propose a special purpose image processing algorithm based on Eigen vector to recognize various signs of Indian Sign Language forlive video sequences with high accuracy. Various difficulties faced by different researchers have been tried to minimize with our approach. Recognition rate of 96.25% was achieved. The experiment was carried out with bare hands, thus removing the difficulty faced using the gloves. We have extended our work [4] for video sequence in this paper.

## 2. ISL AND SYSTEM OVERVIEW
### 2.1 ISL Alphabets
Indian Sign Language was developed so that the deaf people in the society can interact with the normal people without any difficulties. Here in this paper, we have considered the alphabets of ISL which involves the use of either single hand or both hands. A total of 24 alphabets were considered which is shown in Figure 1.

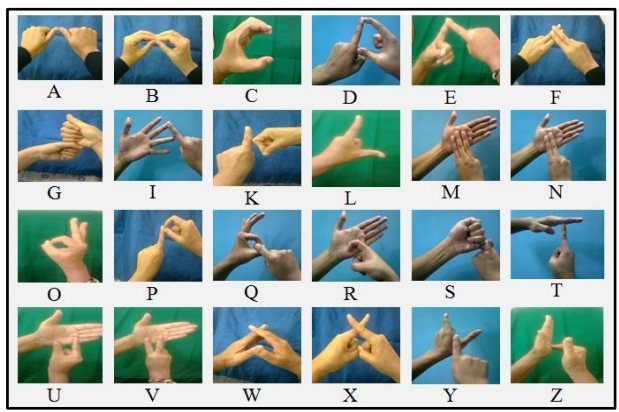

**Fig 1: Various alphabets of Indian Sign Language**

## 3. PREPROCESSING OF SIGN LANGUAGE RECOGNITION

### 3.1 Data Acquisition

The first step for our proposed system is the capturing of the video using webcam where different alphabets were taken into consideration. 24 different alphabets were considered for testing from 20 people. Some of the continuous video frames captured are given in Figure 3.

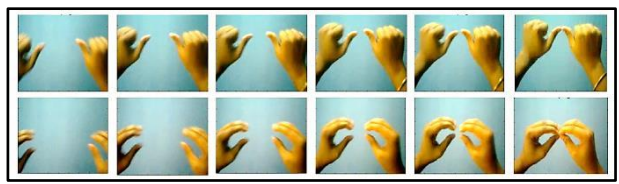

**Fig 3: Some of the video frames captured**

### 3.2 Detection of Hand Gestures

Skin Filtering was performed to the input video frames for detection of hand gestures. It was done so that the required hand could be extracted from the background. Skin Filtering is a technique used for separating the skin colored regions from the non-skin colored regions. The steps used in this skin filtering are shown in Figure 4 as explained in [4, 13].

### 2.2 System Overview

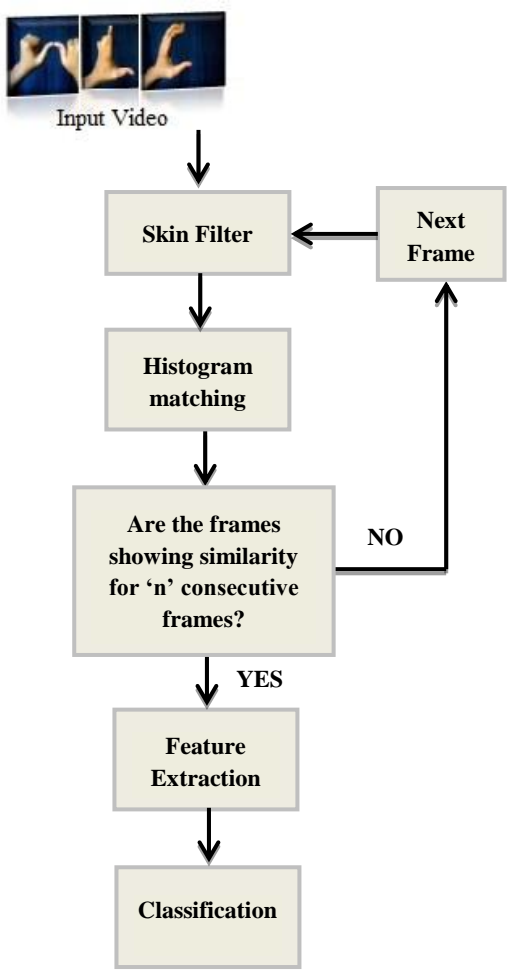

**Fig 2: Overview of the proposed system**

The proposed system is shown in Figure 2 which comprises of 3 major stages-preprocessing stage which includes the skin filtering and histogram matching to find out the similarity between frames, Feature Extraction stage in which the Eigen values and Eigen vector are being considered as features and finally Eigen value weighted Euclidean distance based classification technique as used in [4]. The details of each stage will be discussed in the following sections.

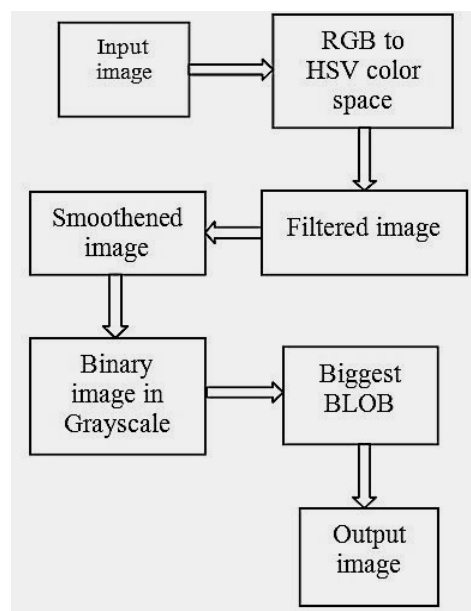

**Fig 4: Block diagram of Skin Filtering**

At first, the input frame was converted to HSV color space. This step was taken because HSV color space was less sensitive to illumination changes compared to RGB. Then it was filtered, smoothened and finally the biggest binary linked object was being considered so as to avoid consideration of skin colored objects other than hand. The resultant image is a binary image with hand regions in white and background in black color. The filtered hand is then found out. The results achieved using skin filtering is given in Figure 5.

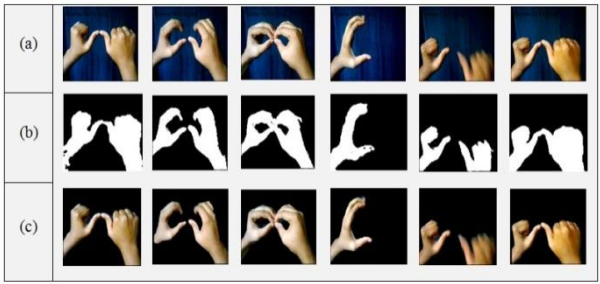

**Fig 5: (a) Input video frames, (b) Skin filtered result, (c) Filtered Hand**

### 3.3 Histogram Matching
After extracting out the skin colored regions from the background, histogram matching is done in the next step. They following steps describe the process of histogram matching:

**Step 1:** The histograms of all the frames of the video are found out.

**Step 2:** The similarities of the consecutive frames are checked by finding out the difference of their histogram.

$$Difference(n) = Hist(n) - Hist(n-1) \quad (1)$$

where Hist represents Histogram and n represents current frame.

**Step 3:** If the difference is found to be above a certain threshold, they are considered as similar. This difference is found out for 'n' number of frames. We have chosen the threshold 'n' to be 17.

**Step 4:** If all the 'n' frames shows similarities, then it is considered to be an unidentified sign and further steps of feature extraction and classification is carried on.

## 4. FEATURE EXTRACTION
Feature Extraction stage is necessary because certain features has to be extracted so that they are unique for each gesture or sign. After the decision is made that a sign is present, then the last frame is taken into consideration and features like Eigen values and Eigen vectors are extracted from that frame. The procedure to calculate the features i.e. Eigen value and Eigen vector are given in as follows:

**Step 1:** Frame resizing- Let us assume the last frame is 'X'. 'X' is resized to 70 by 70.

**Step 2:** Mean and Covariance calculation- Mean 'M' and Covariance 'C' is calculated as given in [1].

$$M = E\{X\} \quad (2)$$
$$C = E\{(X-M)(X-M)'\} \quad (3)$$

**Step 3:** The Eigen values and Eigen vectors are calculated from the above covariance 'C' and the Eigen vectors are arranged in such a manner that the Eigen values are in descending order.

**Step 4:** Data Compression- Out of 70 Eigen vectors only first 5 principle vectors were considered, thus reducing the dimension of the matrix.

## 5. CLASSIFICATION
After the features like Eigen values and Eigen vectors are extracted from the last frame, the next stage is to compare it with the features of the signs already present in the database for classification purpose. It was achieved by considering Eigen value weighted Euclidean Distance based classification Technique as in [4]. The steps for our classification technique are described as follows:

**Step 1:** Calculation of Euclidean Distance- ED was found out between the Eigen vectors calculated from the test frame of the video and the Eigen vectors of the images already present in the database.

$$ED = \sqrt{\sum_{n=1}^{5}(VT(n) - VD(n))^2} \quad (4)$$

Where VT is the Eigen vector of the test frame and VD is the Eigen vector of the database image.

**Step 2:** Calculation of Eigen value difference- The difference between the Eigen value of the database images and theEigen value of the current video frame was found out.

**Step 3:** The above difference was then multiplied with the Euclidean distance obtained.

**Step 4:** After the above operation was carried out, the results obtained for each image was added. After addition, the minimum of all was checked. The minimum represented the recognized symbol.

## 6. EXPERIMENTAL RESULTS AND ANALYSIS
The Indian Sign Language recognition approach was implemented using MATLAB version 7.6 (R2008a) as software and Intel® Pentium® CPU B950 @ 2.10GHz processor machine, Windows 7 Home basic (64 bit), 4GB RAM and awebcam with resolution of 320x240.

### 6.1 Data set and Parameters considered
The data set used for training the recognition system consisted of 24 signs of ISL for 20 people. Thus a total of 480 images were stored in database. We had tested our system with 20 videos and achieved a good success in it.One parameter was considered in our system i.e. the threshold 'n' which is the number of frames it has to check for similarity to determine whether it was a sign or not.

### 6.2 Results and Recognition Rate
Table 1 describes one of the video frame and its results obtained using Eigen value weighted Euclidean distance based classification technique for few images. Similar procedure is carried out for other video frames.The overall recognition rate was calculated and found to be 96.25%. Table 2 describes the success rate for different signs of ISL.

## 7. CONCLUSION AND FUTURE WORK
A fast, novel and robust system was proposed for recognition of different alphabets of Indian Sign Language for video sequences. Skin filtering was used for detection of hands, Eigen vectors and Eigen values were considered as the features and finally effective classification was achieved using Eigen value weighted Euclidean Distance based classifier. Features like good accuracy, use of bare hands, recognition of both single and both hand gestures, working with video were achieved by us when compared to other related works.

**Table 1. Eigen Value Weighted Euclidean Distance Based Classification**

| Current video frame | Database image | Eigen value weighted ED (1st Eigen vector) | Eigen value weighted ED (2nd Eigen vector) | Eigen value weighted ED (3rd Eigen vector) | Eigen value weighted ED (4th Eigen vector) | Eigen value weighted ED (5th Eigen vector) | Sum | Recognized symbol |
|---|---|---|---|---|---|---|---|---|
| 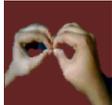 | A | 1.5222 | 5.2195 | 1.7085 | 0.3343 | 0.0008 | 8.7853 | "B" |
| | B | 0.0663 | 0.0852 | 0.9029 | 0.3995 | 0.0713 | **1.5252** | |
| | C | 3.5017 | 0.4849 | 2.4430 | 0.7793 | 0.0005 | 7.2094 | |
| | D | 2.6072 | 0.9669 | 2.0515 | 0.3732 | 0.7585 | 6.7573 | |
| | E | 1.1572 | 1.4187 | 2.6283 | 1.7608 | 0.3713 | 7.3363 | |
| | F | 1.4314 | 0.5373 | 2.1742 | 1.2409 | 0.3484 | 5.7322 | |
| | G | 6.8007 | 1.8617 | 5.1714 | 1.5333 | 0.8005 | 16.1676 | |
| | I | 0.1137 | 1.5154 | 2.6893 | 0.0671 | 0.7337 | 5.1192 | |
| | K | 6.0554 | 5.5240 | 0.6496 | 0.5386 | 0.4044 | 13.172 | |
| | L | 0.8418 | 1.9064 | 2.8906 | 1.8512 | 0.6212 | 8.1112 | |
| | M | 1.0951 | 4.0791 | 0.9159 | 0.2208 | 2.1370 | 8.4479 | |
| | N | 1.3116 | 1.4438 | 1.1714 | 1.6105 | 0.4341 | 5.9714 | |
| | O | 1.1226 | 0.1808 | 2.2131 | 0.1278 | 0.1857 | 3.8300 | |
| | P | 4.0148 | 1.0833 | 0.8182 | 0.5821 | 0.2839 | 6.7823 | |
| | Q | 6.4535 | 4.0007 | 2.0588 | 0.1288 | 0.1052 | 12.747 | |
| | R | 0.7353 | 1.7165 | 1.6045 | 1.0081 | 0.1000 | 5.1644 | |
| | S | 7.8506 | 2.2420 | 0.8310 | 0.5543 | 0.3873 | 11.8652 | |
| | T | 1.3719 | 2.9341 | 2.5987 | 1.6969 | 0.5597 | 9.1613 | |
| | U | 1.0822 | 3.3901 | 2.7790 | 0.5545 | 0.1775 | 7.9833 | |
| | V | 3.2136 | 6.8409 | 3.3667 | 0.7386 | 0.0737 | 14.2335 | |
| | W | 5.3181 | 1.8345 | 0.2591 | 1.3791 | 0.0003 | 8.7911 | |
| | X | 2.4952 | 0.8294 | 3.2854 | 0.3123 | 0.2766 | 7.1989 | |
| | Y | 9.6108 | 1.8892 | 1.9481 | 0.3101 | 0.2358 | 13.994 | |
| | Z | 1.3720 | 0.7915 | 4.3744 | 1.7377 | 0.1960 | 8.4716 | |

**Table 2. Success rates of Classification**

| Symbol | Number of images experimented | Number of correct recognition | Success rate |
|---|---|---|---|
| A | 20 | 18 | 90% |
| B | 20 | 20 | 100% |
| C | 20 | 20 | 100% |
| D | 20 | 20 | 100% |
| E | 20 | 18 | 90% |
| F | 20 | 20 | 100% |
| G | 20 | 20 | 100% |
| I | 20 | 20 | 100% |
| K | 20 | 19 | 95% |
| L | 20 | 20 | 100% |
| M | 20 | 19 | 95% |
| N | 20 | 18 | 90% |
| O | 20 | 20 | 100% |
| P | 20 | 20 | 100% |
| Q | 20 | 20 | 100% |
| R | 20 | 18 | 90% |
| S | 20 | 20 | 100% |
| T | 20 | 20 | 100% |
| U | 20 | 16 | 80% |
| V | 20 | 18 | 90% |
| W | 20 | 20 | 100% |
| X | 20 | 18 | 90% |
| Y | 20 | 20 | 100% |
| Z | 20 | 20 | 100% |

Table 3 describes the comparison of our work with the other related works done.

We have extended our work from static image recognition [4] of ISL to live video recognition. In future, we will try to extend our work in real time with better accuracy. And attempts will be made to extend the work towards more words and sentences.

**Table 3. Comparative study between our work and other approaches**

| Name of the technique used | Success Rate | Remark |
|---|---|---|
| Karhunen-Loeve Transform [1] | 98% | • Recognition of only single hand gestures.<br>• Few hand gestures considered.<br>• Some gestures could not be recognized.<br>• Worked on static images. |
| Eigen value weighted Euclidean distance [4] | 97% | • Recognition of both single and two hand gestures was made possible.<br>• Worked only on static images. |
| Hidden Markov Model [7] | 99% | • Worked on video sequence.<br>• Use of colored gloves- a limitation.<br>• Recognition of American Sign Language with single hand gestures. |
| HMM [9] | 84% | • Though worked on real time but very Low accuracy rate<br>• Use of Data Gloves<br>• Recognition of single hand gestures |
| **Our work** | **96.25%** | • Works on video sequences.<br>• Recognition of both single and two hand gestures.<br>• High accuracy rate in video processing.<br>• Use of bare hands. |

## 8. ACKNOWLEDGMENTS